\documentclass[11pt,a4paper]{article}
\PassOptionsToPackage{breaklinks}{hyperref}
\usepackage[hyperref]{emnlp2020}
\pdfoutput=1
\usepackage{times}
\usepackage{latexsym}

\usepackage{microtype}

\aclfinalcopy % Uncomment this line for the final submission
\title{Manipulating emotions for ground truth emotion analysis}

\author{Bennett Kleinberg \\
  Department of Security and Crime Science \\
  Dawes Centre for Future Crime \\
  University College London \\
  \texttt{bennett.kleinberg@ucl.ac.uk} }

\date{}

\begin{document}
\maketitle
\begin{abstract}
Text data are being used as a lens through which human cognition can be studied at a large scale. Methods like emotion analysis are now in the standard toolkit of computational social scientists but typically rely on third-person annotation with unknown validity. As an alternative, this paper introduces online emotion induction techniques from experimental behavioural research as a method for text-based emotion analysis. Text data were collected from participants who were randomly allocated to a happy, neutral or sad condition. The findings support the mood induction procedure. We then examined how well lexicon approaches can retrieve the induced emotion. All approaches resulted in statistical differences between the true emotion conditions.  Overall, only up to one-third of the variance in emotion was captured by text-based measurements. Pretrained classifiers performed poorly on detecting true emotions. The paper concludes with limitations and suggestions for future research.

\end{abstract}

\section{Introduction}

We leave ever-increasing traces of our behaviour on the Internet, most prominently in the form of text data. People comment on YouTube videos, broadcast their opinions through social media or engage in fringe forum discussions \cite{hine2016a}. An abundance of “text as data” has led to the emergence of research studying human emotion through text \cite{gentzkow2019a}.

\subsection{Detecting emotions in text}
Emotion analysis is an umbrella term for the task aimed at "determining one’s attitude towards a particular target or topic. [Where A]ttitude can mean an evaluative judgment […] or an emotional or affectual attitude such as frustration, joy, anger, sadness, excitement" \cite{mohammad2020a}.  The applications of emotion analysis are wide-ranging from business use cases (e.g. understanding how consumers feel about a product) to public health issues. For example, researchers have used text data to understand affective responses to COVID-19 \cite{li2020a, vegt2020a}, and are seeking to harness further text data to study mental health aspects and coping mechanisms to a new normal.

Researchers typically resort to one of two options to learn about human emotions from text: lexicon approaches or (pretrained) classification models. A lexicon is typically a list of \textit{n-}grams that are assigned a sentiment (or emotion) value – either through human crowdsourcing effort \cite{mohammad2013a, mohammad2010a, socher2013a, ding2016a}  or (semi) automated means such as hashtags of Tweets \cite{mohammad2015a, sailunaz2019a}. The outcome value of a document or sentence is often represented as an aggregate sentiment or emotion score of the units of analysis in the document \cite[for an overview see][]{poria2020a}.

Sentiment classification typically relies on supervised machine learning. Here, classification algorithms trained on annotated corpora make predictions about the overall sentiment of an unseen, new document (e.g. positive, neutral, negative). Importantly, to train classifiers, researchers here too need labelled data generated through either manual or semi-automated efforts \cite[for another comprehensive overview on emotion analysis, see][]{mohammad2020a}.

\subsection{Text as a proxy of human emotion}

While most emotion analysis studies focus on predicting sentiment, relatively little attention is paid to how we arrive at labelled data in the first place. We argue that as a research community, we are only marginally interested in the sentiment or emotion of a \textit{document}. Instead, the primary objective we have with emotion analysis is to make inferences about the emotional state of the \textit{author}. Consequently, text data are only a convenient proxy for emotional processes. The latter are typically inaccessible from observational data (e.g. Twitter data), hence we use text data as a backdoor through which emotion can be studied. 

Put differently, we care only about text data to the degree to which they allow us to make inferences about human emotions. In order to test how well computational linguistics approaches are capable of making these inferences, it is worthwhile looking at various types of ground truth. These are, in turn, intertwined with the core challenge for emotion analysis, that emotions are inherently subjective.

\subsection{Degrees of ground truth}

Emotion analyses typically require data labelled concerning an emotion of interest (e.g., happiness, sadness, anger). The procedures applied to obtain these data result in three degrees of ground truth with implied degrees of label validity. The widely used approach of using readily-available, observational data and imposing a third-person label on the retrieved text data is what we call \textit{pseudo ground truth}. The label validity is unknown because we cannot examine how the emotion judgment of a third person aligns with the true emotion of the author of the document. For example, just because we label a Tweet as "angry" does not mean that the person who wrote it was indeed angry - although the latter and not the former is the target of study.

To improve the validity of the labels, some studies asked participants to recall and write about an event in which they felt certain emotions - e.g. the ISEAR dataset \cite{seyeditabari2018a} - or asked them directly about their feelings before they wrote a text \cite{kleinberg2020a}. While these efforts increase the confidence in the label validity, they are still vulnerable to the limitations of survey designs such as self-selection (e.g. some people may be more inclined to report positive emotions than others), an accurate recall of events (e.g. people forget or amend a memory), and demand characteristics (e.g. writing an expected or socially desirable manner).

To fully assess the validity of emotion analysis, we thus need stricter label ground truth. The strictest criterion is that the researcher has actively manipulated an outcome through random assignment (e.g. making one group like a product and another group dislike it before they both write a review). By experimentally manipulating the emotional state so that one group of participants feels positive and the other negative, we can be confident the groups only differed in the emotional state. Hence, it allows us to isolate the effect that emotional states have on text data. In this paper, we adopt emotion induction procedures from experimental psychology for brief online settings to conduct the first test of emotion analysis on manipulated ground truth emotions.

\subsection{Experimental emotion induction}

Videos, affirmative statements, and music are but a few stimuli that can be used to induce a specific mood in a person \cite{scott2018a, westermann1996a, heene2007a}. Recently, mood induction studies have started to examine the validity in online settings \cite{marcusson-clavertz2019a}. The authors measured the positive and negative affect of participants before and after an emotion manipulation: participants were shown \textit{(i)} a happy video and listened to uplifting music, \textit{(ii)} a sad video and sad music, or \textit{(iii)} a neutral video and music. The procedure was successful in inducing the desired emotion (joviality in the happy emotion induction, and sadness in the sad emotion induction).

Suchlike procedures have not yet been used in NLP emotion analysis research. A noteworthy exception is the 2014 study on emotional contagion on Facebook \cite{kramer2014a}. The researchers manipulated the exposure to positive and negative words in users' Facebook news feeds. When a user saw fewer positive words, they tended also to produce fewer positive and more negative words themselves; and vice versa. To the best of our knowledge, this is the only attempt to experimentally manipulate the mood of participants and assess how this is reflected in subsequent language behaviour. 

\subsection{Aims of this paper}

In the current paper, we combine active mood induction manipulation and language measurements in a behavioural experiment to assess the validity of emotion analysis approaches.

\section{Method and Data}

The local ethics committee has approved this study.\footnote{The data are available at \url{https://osf.io/nxr2a/?view_only=441e160f2ab8490e9262c5f390114bb5}.}

\subsection{Participants}

We collected data of 525 participants through the crowd-sourcing platform Prolific Academic.\footnote{The sample size was determined a priori through power analysis for a small-to-moderate effect size (Cohen’s $f = 0.20$), a significance level of 0.01, and a statistical power of 0.95.} One participant was excluded because the text was written in Spanish, and seven others were excluded because their text input contained more than 20\% punctuation. The final sample of 517 participants ($\mathrm{M_{age}}=32.11$ years,  $\mathrm{SD}=11.84$; 45.47\% female) were randomly allocated to the happy mood condition ($n=175$), sad mood condition ($n=169$), or the neutral condition ($n=173$). Neither age nor gender differed statistically between the three groups.

\subsection{Experimental task}

All participants accessed the experimental task online, provided informed consented and were told that this study was about their feelings at this very moment. Each participant was randomly allocated to one of three experimental conditions (see \ref{Emotion manipulation}). In each condition, the participants watched a movie clip and made a self-assessment of their current feelings on a 9-point scale (1 = not at all, 5 = moderately, 9 = very much). They specified how much happiness, sadness, anger, disgust, fear, anxiety, relaxation, and desire they felt at that moment \cite{harmon-jones2016a}. Also, they chose which of these eight emotions best characterized their current feeling if they had to choose just one.

Upon completing these questions, all participants were asked to write a few sentences (at least 500 characters) to \textit{"express [their] feelings at this moment"}. After that, all participants were asked to rate how well they think they could express their feelings in the text (on a 9-point scale: 1 = not at all, 5 = moderately, 9 = very well) and received a debriefing. During the debriefing, the participants were given a chance (regardless of their condition) to watch the happy movie. This ensured that no participant left the task in a sad mood.

\subsection{Emotion manipulation}
\label{Emotion manipulation}

We adapted the mood induction procedure from previous work \cite{marcusson-clavertz2019a}.

The negative mood induction procedure consisted of a video from The Lion King, showing the young lion Simba in danger in a stampede. Simba's father is rushing to his rescue but dies after being run over by wildebeest. The scene ends with Simba searching for his father only to find his lifeless body. 

The positive mood induction consisted of a video clip, again from the Lion King, showing the characters Timon and Pumba singing the song Hakuna Matata. 

In the neutral condition, the participants watched a video about magnets from the documentary series Modern Marvels. Each video clip was edited to two minutes in duration.

\begin{table*}[h!]
\centering
\begin{tabular}{lccc}
\hline \textbf{} & \textbf{Positive emotion} & \textbf{Neutral} & \textbf{Negative emotion} \\ \hline
Text length          & 119.95 (21.93) & 122.65 (24.41) & 122.04 (31.01) \\
\textit{Self-reported emotion values}     &               &               &               \\
Happiness            & 6.06 (2.04)    & 4.75 (1.97)    & 2.87 (1.95)    \\
Sadness              & 2.71 (1.99)    & 3.28 (2.44)    & 6.37 (2.52)    \\
\textit{Sentiment values}     &               &               &               \\
Jockers              & 1.36 (3.45)    & 0.43 (3.54)    & -0.95 (2.82)   \\
NRC                  & 1.14 (3.87)    & 0.51 (3.87)    & -0.28 (3.44)   \\
Bing                 & 0.22 (4.37)    & -0.58 (4.34)   & -2.70 (3.48)   \\
Afinn                & 2.34 (8.97)    & -0.13 (8.97)   & -3.88 (7.55)   \\
SenticNet 4          & 0.10 (0.25)    & 0.05 (0.23)    & -0.04 (0.24)   \\
LIWC Tone            & 51.80 (38.88)  & 40.18 (37.73)  & 18.64 (27.51) \\
\hline
\end{tabular}
\caption{\label{Table 1} Descriptive statistics ($M$, $SD$) per condition.}
\end{table*}

\subsection{Lexicon measurements}

We obtained the mean sentiment value for each text from six lexicons: (1) Matthew Jockers sentiment lexicon of 10,738 unigrams rated between -1.00 and +1.00 \cite{jockers2015a}, (2) the NRC sentiment lexicon \cite{mohammad2010a} consisting of 13,891 terms associated with eight emotions and positive/negative sentiment (here we only used the sentiment terms), (3) Bing Liu’s sentiment lexicon (6,789 unigrams scored as -1 or  +1), and (4) the AFINN sentiment lexicon \cite{nielsen2011a} of 2,477 unigrams scored from -5 to +5, (5), the SenticNet 4 lexicon (23,626 terms scored between -0.98 and +0.98)\cite{cambria2016a} (6) the LIWC category “Tone” which represents the proportion of words in a target text that belong to the emotional tone lexicon \cite{pennebaker2015a}.

\subsection{Pretrained classifiers}

We used three pretrained sentiment classifiers which predict positive vs negative sentiment trained on the Stanford Sentiment Treebase 2 (SST2, \cite{socher2013a}, and the IMDB Movie Review dataset \cite{maas_learning_2011}. The models were identical to the baselines used in earlier work and and include a CNN, LSTM and BERT model \cite[for details, see][]{mozes_frequency-guided_2020}. The baseline accuracy rates reached with theses models in their respective dataset are: $CNN_{imdb} = 0.87$, $LSTM_{imdb} = 0.87$, $BERT_{imdb} = 0.91$ and $CNN_{sst2} = 0.84$, $LSTM_{sst2} = 0.84$, $BERT_{sst2} = 0.92$ \cite{mozes_frequency-guided_2020}.

\subsection{Analysis plan}

First, we examine whether the emotion induction manipulation was successful. Second, we investigate whether the mean sentiment scores of the texts differed per induced emotion. Third, we examine how much of the self-reported mood scores are captured by each measurement. Lastly, we assess how well pretrained models detect the induced emotions.

\section{Results}

For the statistical analyses, we report effect sizes from null hypothesis significance testing. The reporting of p-values is now widely discouraged due to their misinterpretation and bias for large sample sizes \cite{benjamin_redefine_2018}. Effect sizes provide a standardised measure of the magnitude of a phenomenon and are comparable across studies \cite{lakens_calculating_2013}.

The effect size Cohen's $d$ expresses the mean difference between two groups divided by the pooled standard deviation.\footnote{For interpretation: a $d$ of 0.2, 0.5, and 0.8 represent a small effect, medium-sized and large effect, respectively \cite{cohen1988a}}. Squared [brackets] denote the 99\% confidence interval of the effect size.

\subsection{Emotion manipulation}

Self-reported happiness: There was a significant effect of emotion manipulation on the self-reported happiness. The happiness score in the happy condition was higher than in the neutral, $d=0.65$ $[0.37; 0.93]$; and the sad condition, $d=1.60$ $[1.28; 1.91]$. The happiness score in the neutral condition, in turn, was higher than that in the sad condition, $d=0.96$ $[0.66; 1.25]$ (Table \ref{Table 1}).

Self-reported sadness: Similarly, a significant effect of emotion manipulation on the self-reported sadness, showed that the sadness score in the sad condition was higher than in the neutral, $d=1.24$ $[0.94; 1.55]$; and the happy condition, $d=1.61$ $[1.29; 1.92]$. There was no difference in the sadness score between the neutral and happy condition, $d=0.26 [0.02; 0.53]$. 
These findings provide substantial evidence for a successful emotion induction.

\begin{table*}[h!]
\centering
\begin{tabular}{lrrr}
\hline \textbf{Measurement} & \textbf{Happy vs Sad} & \textbf{Happy vs Neutral} & \textbf{Neutral vs Sad} \\ \hline
Jockers     & 0.74* {[}0.45; 1.03{]} & 0.27 {[}-0.01; 0.55{]} & 0.43* {[}0.15; 0.72{]}     \\
NRC         & 0.39* {[}0.11, 0.67{]}           & 0.16 {[}-0.11, 0.44{]} & 0.22 {[}-0.06; 0.49{]}           \\
Bing        & 0.74* {[}0.43; 1.03{]} & 0.19 {[}-0.09; 0.46{]} & 0.54* {[}0.26; 0.82{]} \\
Afinn       & 0.75* {[}0.46; 1.04{]}  & 0.28 {[}0.00; 0.55{]} & 0.45* {[}0.17; 0.74{]}         \\
SenticNet 4 & 0.56* {[}0.27; 0.84{]}  & 0.21 {[}-0.06; 0.49{]} & 0.36* {[}0.08; 0.64{]}          \\
LIWC “Tone” & 0.99* {[}0.69; 1.28{]} & 0.30 {[}0.03; 0.58{]}  & 0.65* {[}0.37; 0.94{]}\\
\hline
\end{tabular}
\caption{\label{Table 2} Statistical comparisons of lexicon measurements per condition.}
\end{table*}

\subsection{Emotion measurements}

The texts written by the participants were, on average, 121.54 words long ($SD = 25.99$). The average length did not differ between the three conditions (Table \ref{Table 1}).

The results for the overall manipulation and the follow-up comparisons (Table \ref{Table 2}) suggest that all six sentiment measures differed across all three groups with the most substantial effect for the LIWC Tone measurement.\footnote{ * = Cohen’s $d$ sign. at $p < .001$. For all variables, there was a significant main effect of the emotion manipulation.} All measurements further found substantial differences between the happy and sad condition.

Texts in the happy condition were substantially more positive than in the sad condition on all six linguistic measurements. The largest difference was found for the LIWC Tone measurement ($d=0.99$) while the smallest was evident for the NRC approach ($d=0.39$). In none of the happy vs neutral comparisons was there a notable statistical difference. Except for the NRC approach, there were higher sentiment scores in the neutral than in the sad condition, with the LIWC Tone measurement showing the largest effect ($d=0.65$).

\begin{table}
\centering
\begin{tabular}{lrr}
\hline \textbf{} & \textbf{Happiness} & \textbf{Sadness}  \\ \hline
Jockers     & 28.94     & 21.81   \\
NRC         & 15.52     & 10.63   \\
Bing        & 26.83     & 22.94   \\
Afinn       & 28.72     & 23.72   \\
SenticNet 4 & 27.65     & 15.44   \\
LIWC “Tone” & 30.76     & 25.91 \\
\hline
\end{tabular}
\caption{\label{Table 3} $R^2$ (in \%) between reported emotion values and linguistic measurement}
\end{table} 

\subsection{Correlation between induced emotion and measurement}

Correlational analysis shows that the emotion measures are all significantly positively correlated to the happiness scores and negatively correlated to the sadness scores.\footnote{The Pearson $r$ values can be obtained by taking the square root of the reported $R^2$} We use the $R^2$ to express the percentage of variance in happiness and sadness values that is explained by either emotion measurement (Table \ref{Table 3}).

The emotion measurements indicate similar results; all explain between 26.83 and 30.76\% of the variance in the happiness scores and 21.94 to 25.91\% of the sadness score variance. Only the NRC deviates with 15.52\% and 10.63\% of the happiness and sadness scores being explained, respectively. That is, up to one-third of the self-reported happiness and just above one quarter of the self-reported sadness are explained by linguistic measurements.

\subsection{Predicting emotions}

Table \ref{Table 4} shows the area under the curve results for the pretrained models and the lexicon approaches.\footnote{Read the pretrained results as follows: "SST2: CNN" means that the classifier was a conv. neutral netword trained on the SST2 corpus} For the latter, the AUCs ranged from 0.60 (NRC) to 0.70 (Jockers, Bing, Afinn) and 0.74 (LIWC). None of the pretrained models was able to obtain higher AUCs than the lexicon approaches. For the pretrained models we also calculated the prediction accuracy. None of the models outperformed the chance level significantly.

\begin{table}
\centering
\begin{tabular}{lcc}
\hline \textbf{Approach} & \textbf{AUC} & \textbf{Acc.}  \\ \hline
\textit{Lexicon approaches} &                      \\
Jockers            & 0.70 {[}0.63; 0.77{]} & - \\
NRC                & 0.60 {[}0.53; 0.68{]} & - \\
Bing               & 0.70 {[}0.63; 0.78{]} & - \\
Afinn              & 0.70 {[}0.63; 0.77{]} & - \\
SenticNet 4        & 0.67 {[}0.60; 0.75{]} & - \\
LIWC “Tone”        & 0.74 {[}0.67; 0.81{]} & - \\
\textit{Pretrained models}  &                      \\
SST2: CNN          & 0.65 {[}0.57; 0.72{]} & 0.62 \\
SST2: LSTM         & 0.63 {[}0.55; 0.70{]} & 0.60 \\
SST2: BERT         & 0.63 {[}0.55; 0.71{]} & 0.60 \\
IMDB: CNN          & 0.59 {[}0.61; 0.67{]} & 0.54 \\
IMDB: LSTM         & 0.56 {[}0.48; 0.64{]} & 0.54 \\
IMDB: BERT         & 0.63 {[}0.55; 0.71{]} & 0.62 \\
\hline
\end{tabular}
\caption{\label{Table 4} Area under the curve [95\% CI] and accuracy for predicting the pos-vs-neg emotion induction.}
\end{table}

\subsection{Moderation through language proficiency}

A factor that could have influence the current findings is the participants' native language (English vs not English) and their self-reported ability to express their feeling in the text. We added both factors as covariates to the statistical models for each linguistic emotion measurement. There was no evidence for an effect of these and conclude that these factors did not influence the effect of the emotion manipulation.

\section{Discussion}

Emotion analysis is used widely and an evaluation with experimentally manipulated ground truth data can help study emotions with higher label validity. We randomly assigned participants to a happy, neutral or sad mood. Our findings suggest \textit{i)} that mood induction can be done within a concise time online making it a method usable in larger scale settings; \textit{ii)} that frequently used emotion measurements only capture a small portion of the happiness and sadness of the authors; and \textit{iii)} that pretrained models perform poorly on predicting ground truth emotion data.

\subsection{Manipulating emotions online}

By randomly assigning participants to the happy, sad or neutral condition, we were able to attribute any change in the reported emotion to the experimental manipulation. There were substantial statistical differences in happiness and sadness when our participants were in the respective experimental conditions. While emotion (or mood) induction in itself is not new in experimental behavioural research \cite{westermann1996a, marcusson-clavertz2019a}, our work tested this method for text-based analyses and demonstrated its feasibility within short time online.  The current procedure could thus serve as an introduction of an experimental emotion induction approach for NLP research.

\subsection{Inferring ground truth emotions}

\subsubsection{Lexicon approaches}
This paper assessed how a range of emotion analysis methods - lexicons and pretrained classifiers - could infer the true emotion of an author. For all of the examined lexicon measures, large statistical differences emerged between the texts written by participants in the happy and the sad condition. The findings varied, however, by the approach used: while the Jockers, Bing and Afinn lexicons resulted in moderate to large effect sizes, SenticNet4 and the NRC approach showed only moderate differences. Using the LIWC approach yielded the largest difference. 

Taken together, these findings indicate that the emotion induction resulted in texts that differed in the expected directions on sentiment (i.e. happy participants wrote positive texts and sad participants negative texts). Crucially, the randomised experimental design allows us to conclude that the textual differences can be attributed to the difference in emotion - which classical observational studies cannot.

\subsubsection{Prediction with pretrained models}
Lexicon approaches are only one means to make inferences about emotions. Despite their widespread use \cite[e.g.][]{kramer2012a, guillory2011a}, machine learning classifiers typically beat lexicon approaches in accuracy.\footnote{See \url{https://github.com/sebastianruder/NLP-progress/blob/master/english/sentiment_analysis.md}} We therefore also tested three models trained on vast sentiment corpora. 

Although the models stem from a different domain (movie reviews), we would have expected these models to perform well on making a simple positive (happy) vs negative (sad) prediction. Surprisingly, none of the AUCs of the pretrained models outperformed the lexicon approaches. Neither did any of the pretrained models exceed random guessing performance.

\subsubsection{Explained variance in emotions}
Another aspect is how much of the emotion is captured by either linguistic measure. The emotion measurement explained between 27\% and 31\% of the happiness scores (except 16\% for the NRC approach) and 22\% and 26\% of the sadness scores (except for 11\% and 15\% for NRC and SentiNet4) . If 31\% of the variance of happiness are explained in emotion analysis, 69\% is not. That implies that more than two-thirds of the emotion of interest is not captured.

\subsection{Limitations and outlook}

Our work comes with a few limitations. First, one might argue that models built in the current domain would inevitably have performed better than the pretrained models from the movie reviews domain. We argue that most applied emotion analysis problems are unsupervised problems and hence inevitably rely on pretrained models or lexicons. For example, if a clinician wants to obtain insights into a patient's mental health (e.g., based on diaries), training models would defeat the purpose. Oftentimes there simply is no ground truth and we have to rely on measurements.

Second, the emotion induction procedure used in the current study is a simple happiness manipulation. As such, it is likely an oversimplification of human emotions which can come in mixtures \cite[e.g. feeling sad and happy at the same time, ][]{larsen_can_2001}. To date, manipulation procedures exist beyond the happiness-sadness dichotomy \cite[e.g. making people angry, see][]{lobbestael_how_2008} and meta-analytical research suggests even mixed emotions can be induced experimentally \cite{berrios_eliciting_2015}. However, these are not yet usable in an online setting. Future work on enabling complex emotion induction online is needed to apply the current approach to more complex emotions.

A central question for future work is not only how emotion detection can be improved to capture true emotional states better, but also, to understand the process and other sources that explain emotions. For example, it is possible that language proficiency - or even more specifically, a skill to express emotions in writing - moderates one's ability to express their feelings in the form of text. Formalising the path from emotion-to-text-to-measurement that is underlying so many socio- and psycholinguistic approaches, would allow us to map out the boundaries of emotion analysis and make the next steps in the field. Ultimately, deepening the understanding of how emotions manifest themselves in text, and how we can use computational methods to infer the emotion contained in text, is essential to the study of human cognition and behaviour through text data.

\section{Conclusion}

The current work provided the first insights into experimental emotion induction in computational linguistics research. 

We found that linguistic measurements only partially explain the true emotions. Understanding the nexus from cognition to language is one of the fundamental challenges for computational research about human behaviour. The current paper aimed to contribute to that task by demonstrating the experimental method for emotion manipulation.

\section*{Acknowledgments}
Special thanks to David Marcusson-Clavertz for sharing the mood induction material of his work.

\bibliography{emnlp2020}

\begin{thebibliography}{34}
\expandafter\ifx\csname natexlab\endcsname\relax\def\natexlab#1{#1}\fi

\bibitem[{Benjamin et~al.(2018)Benjamin, Berger, Johannesson, Nosek,
  Wagenmakers, Berk, Bollen, Brembs, Brown, Camerer, Cesarini, Chambers, Clyde,
  Cook, Boeck, Dienes, Dreber, Easwaran, Efferson, Fehr, Fidler, Field,
  Forster, George, Gonzalez, Goodman, Green, Green, Greenwald, Hadfield,
  Hedges, Held, Ho, Hoijtink, Hruschka, Imai, Imbens, Ioannidis, Jeon, Jones,
  Kirchler, Laibson, List, Little, Lupia, Machery, Maxwell, McCarthy, Moore,
  Morgan, Munafó, Nakagawa, Nyhan, Parker, Pericchi, Perugini, Rouder,
  Rousseau, Savalei, Schönbrodt, Sellke, Sinclair, Tingley, Zandt, Vazire,
  Watts, Winship, Wolpert, Xie, Young, Zinman, and
  Johnson}]{benjamin_redefine_2018}
Daniel~J. Benjamin, James~O. Berger, Magnus Johannesson, Brian~A. Nosek, E.-J.
  Wagenmakers, Richard Berk, Kenneth~A. Bollen, Björn Brembs, Lawrence Brown,
  Colin Camerer, David Cesarini, Christopher~D. Chambers, Merlise Clyde,
  Thomas~D. Cook, Paul~De Boeck, Zoltan Dienes, Anna Dreber, Kenny Easwaran,
  Charles Efferson, Ernst Fehr, Fiona Fidler, Andy~P. Field, Malcolm Forster,
  Edward~I. George, Richard Gonzalez, Steven Goodman, Edwin Green, Donald~P.
  Green, Anthony~G. Greenwald, Jarrod~D. Hadfield, Larry~V. Hedges, Leonhard
  Held, Teck~Hua Ho, Herbert Hoijtink, Daniel~J. Hruschka, Kosuke Imai, Guido
  Imbens, John P.~A. Ioannidis, Minjeong Jeon, James~Holland Jones, Michael
  Kirchler, David Laibson, John List, Roderick Little, Arthur Lupia, Edouard
  Machery, Scott~E. Maxwell, Michael McCarthy, Don~A. Moore, Stephen~L. Morgan,
  Marcus Munafó, Shinichi Nakagawa, Brendan Nyhan, Timothy~H. Parker, Luis
  Pericchi, Marco Perugini, Jeff Rouder, Judith Rousseau, Victoria Savalei,
  Felix~D. Schönbrodt, Thomas Sellke, Betsy Sinclair, Dustin Tingley,
  Trisha~Van Zandt, Simine Vazire, Duncan~J. Watts, Christopher Winship,
  Robert~L. Wolpert, Yu~Xie, Cristobal Young, Jonathan Zinman, and Valen~E.
  Johnson. 2018.
\newblock \href {https://doi.org/10.1038/s41562-017-0189-z} {Redefine
  statistical significance}.
\newblock \emph{Nature Human Behaviour}, 2(1):6.

\bibitem[{Berrios et~al.(2015)Berrios, Totterdell, and
  Kellett}]{berrios_eliciting_2015}
Raul Berrios, Peter Totterdell, and Stephen Kellett. 2015.
\newblock \href {https://doi.org/10.3389/fpsyg.2015.00428} {Eliciting mixed
  emotions: a meta-analysis comparing models, types, and measures}.
\newblock \emph{Frontiers in Psychology}, 6.

\bibitem[{Cambria et~al.(2016)Cambria, Poria, Bajpai, and
  Schuller}]{cambria2016a}
Erik Cambria, Soujanya Poria, Rajiv Bajpai, and Bjoern Schuller. 2016.
\newblock Senticnet 4: A semantic resource for sentiment analysis based on
  conceptual primitives.
\newblock In \emph{In}, pages 12,. Osaka, Japan.

\bibitem[{Cohen(1988)}]{cohen1988a}
J.~Cohen. 1988.
\newblock \emph{Statistical power analysis for the behavioral sciences}.
\newblock Academic Press, New York, NY.

\bibitem[{Ding and Pan(2016)}]{ding2016a}
Tao Ding and Shimei Pan. 2016.
\newblock An empirical study of the effectiveness of using sentiment analysis
  tools for opinion mining.
\newblock In \emph{Proceedings of the 12th International Conference on Web
  Information Systems and Technologies}, pages 53–62,, Rome, Italy.
  SCITEPRESS - Science and and Technology Publications.

\bibitem[{Gentzkow et~al.(2019)Gentzkow, Kelly, and Taddy}]{gentzkow2019a}
Matthew Gentzkow, Bryan Kelly, and Matt Taddy. 2019.
\newblock Text as data.
\newblock \emph{Journal of Economic Literature}, 57(3).

\bibitem[{Guillory et~al.(2011)Guillory, Spiegel, Drislane, Weiss, Donner, and
  Hancock}]{guillory2011a}
Jamie Guillory, Jason Spiegel, Molly Drislane, Benjamin Weiss, Walter Donner,
  and Jeffrey Hancock. 2011.
\newblock Upset now?: emotion contagion in distributed groups.
\newblock In \emph{Proceedings of the 2011 annual conference on Human factors
  in computing systems - CHI ’11}, pages 745,, Vancouver, BC, Canada. ACM
  Press.

\bibitem[{Harmon-Jones et~al.(2016)Harmon-Jones, Bastian, and
  Harmon-Jones}]{harmon-jones2016a}
Cindy Harmon-Jones, Brock Bastian, and Eddie Harmon-Jones. 2016.
\newblock The discrete emotions questionnaire: A new tool for measuring state
  self-reported emotions.
\newblock \emph{PLOS ONE}, 11(8).

\bibitem[{Heene et~al.(2007)Heene, Raedt, Buysse, and Oost}]{heene2007a}
Els Heene, Rudi~De Raedt, Ann Buysse, and Paulette~Van Oost. 2007.
\newblock Does negative mood influence self-report assessment of individual and
  relational measures?
\newblock \emph{An Experimental Analysis. Assessment}, 14(1).

\bibitem[{Hine et~al.(2016)Hine, Onaolapo, Cristofaro, Kourtellis, Leontiadis,
  Samaras, Stringhini, and Blackburn}]{hine2016a}
Gabriel~Emile Hine, Jeremiah Onaolapo, Emiliano~De Cristofaro, Nicolas
  Kourtellis, Ilias Leontiadis, Riginos Samaras, Gianluca Stringhini, and
  Jeremy Blackburn. 2016.
\newblock Kek, cucks, and god emperor trump: A measurement study of 4chan’s
  politically incorrect forum and its effects on the web.
\newblock In \emph{Association for the Advancement of Artificial Intelligence}.
\newblock ArXiv: 1610.03452.

\bibitem[{Jockers(2015)}]{jockers2015a}
Matthew Jockers. 2015.
\newblock Syuzhet: Extract sentiment and plot arcs from text.

\bibitem[{Kleinberg et~al.(2020)Kleinberg, van~der Vegt, and
  Mozes}]{kleinberg2020a}
Bennett Kleinberg, Isabelle van~der Vegt, and Maximilian Mozes. 2020.
\newblock Measuring emotions in the covid-19 real world worry dataset.
\newblock ArXiv:2004.04225 [cs], April. arXiv: 2004.04225.

\bibitem[{Kramer(2012)}]{kramer2012a}
Adam~D.I. Kramer. 2012.
\newblock The spread of emotion via facebook.
\newblock In \emph{Proceedings of the 2012 ACM annual conference on Human
  Factors in Computing Systems - CHI ’12}, pages 767,, Austin, Texas, USA.
  ACM Press.

\bibitem[{Kramer et~al.(2014)Kramer, Guillory, and Hancock}]{kramer2014a}
A.D.I. Kramer, J.E. Guillory, and J.T. Hancock. 2014.
\newblock Experimental evidence of massive-scale emotional contagion through
  social networks.
\newblock \emph{Proceedings of the National Academy of Sciences}, 111(24).

\bibitem[{Lakens(2013)}]{lakens_calculating_2013}
Daniël Lakens. 2013.
\newblock \href {https://doi.org/10.3389/fpsyg.2013.00863} {Calculating and
  reporting effect sizes to facilitate cumulative science: a practical primer
  for t-tests and {ANOVAs}}.
\newblock \emph{Frontiers in Psychology}, 4.

\bibitem[{Larsen et~al.(2001)Larsen, McGraw, and Cacioppo}]{larsen_can_2001}
Jeff~T. Larsen, A.~Peter McGraw, and John~T. Cacioppo. 2001.
\newblock \href {https://doi.org/10.1037/0022-3514.81.4.684} {Can people feel
  happy and sad at the same time?}
\newblock \emph{Journal of Personality and Social Psychology}, 81(4):684--696.

\bibitem[{Li et~al.(2020)Li, Li, Li, Alvarez-Napagao, and Garcia}]{li2020a}
Irene Li, Yixin Li, Tianxiao Li, Sergio Alvarez-Napagao, and Dario Garcia.
  2020.
\newblock What are we depressed about when we talk about covid19: Mental health
  analysis on tweets using natural language processing.
\newblock ArXiv:2004.10899 [cs], April. arXiv: 2004.10899.

\bibitem[{Lobbestael et~al.(2008)Lobbestael, Arntz, and
  Wiers}]{lobbestael_how_2008}
Jill Lobbestael, Arnoud Arntz, and Reinout~W. Wiers. 2008.
\newblock \href {https://doi.org/10.1080/02699930701438285} {How to push
  someone's buttons: {A} comparison of four anger-induction methods}.
\newblock \emph{Cognition \& Emotion}, 22(2):353--373.

\bibitem[{Maas et~al.(2011)Maas, Daly, Pham, Huang, Ng, and
  Potts}]{maas_learning_2011}
Andrew~L. Maas, Raymond~E. Daly, Peter~T. Pham, Dan Huang, Andrew~Y. Ng, and
  Christopher Potts. 2011.
\newblock \href {https://www.aclweb.org/anthology/P11-1015} {Learning {Word}
  {Vectors} for {Sentiment} {Analysis}}.
\newblock In \emph{Proceedings of the 49th {Annual} {Meeting} of the
  {Association} for {Computational} {Linguistics}: {Human} {Language}
  {Technologies}}, pages 142--150, Portland, Oregon, USA. Association for
  Computational Linguistics.

\bibitem[{Marcusson-Clavertz et~al.(2019)Marcusson-Clavertz, Kjell, Persson,
  and Cardeña}]{marcusson-clavertz2019a}
David Marcusson-Clavertz, Oscar~N.E. Kjell, Stefan~D. Persson, and Etzel
  Cardeña. 2019.
\newblock Online validation of combined mood induction procedures.
\newblock \emph{PLOS ONE}, 14(6).

\bibitem[{Mohammad and Turney(2010)}]{mohammad2010a}
Saif Mohammad and Peter Turney. 2010.
\newblock Emotions evoked by common words and phrases: Using mechanical turk to
  create an emotion lexicon.
\newblock In \emph{Proceedings of the NAACL HLT 2010 Workshop on Computational
  Approaches to Analysis and Generation of Emotion in Text}, pages 26–34,,
  Los Angeles, CA. Association for Computational Linguistics.

\bibitem[{Mohammad(2020)}]{mohammad2020a}
Saif~M. Mohammad. 2020.
\newblock Sentiment analysis: Detecting valence, emotions, and other affectual
  states from text.
\newblock ArXiv:2005.11882 [cs], May. arXiv: 2005.11882.

\bibitem[{Mohammad and Kiritchenko(2015)}]{mohammad2015a}
Saif~M. Mohammad and Svetlana Kiritchenko. 2015.
\newblock Using hashtags to capture fine emotion categories from tweets.
\newblock \emph{Computational Intelligence}, 31(2).

\bibitem[{Mohammad et~al.(2013)Mohammad, Kiritchenko, and Zhu}]{mohammad2013a}
Saif~M. Mohammad, Svetlana Kiritchenko, and Xiaodan Zhu. 2013.
\newblock Nrc-canada: Building the state-of-the-art in sentiment analysis of
  tweets.
\newblock ArXiv:1308.6242 [cs], August. arXiv: 1308.6242.

\bibitem[{Mozes et~al.(2020)Mozes, Stenetorp, Kleinberg, and
  Griffin}]{mozes_frequency-guided_2020}
Maximilian Mozes, Pontus Stenetorp, Bennett Kleinberg, and Lewis~D. Griffin.
  2020.
\newblock \href {http://arxiv.org/abs/2004.05887} {Frequency-{Guided} {Word}
  {Substitutions} for {Detecting} {Textual} {Adversarial} {Examples}}.
\newblock \emph{arXiv:2004.05887 [cs]}.
\newblock ArXiv: 2004.05887.

\bibitem[{Nielsen(2011)}]{nielsen2011a}
Finn~Årup Nielsen. 2011.
\newblock A new anew: Evaluation of a word list for sentiment analysis in
  microblogs.
\newblock ArXiv:1103.2903 [cs], March. arXiv: 1103.2903.

\bibitem[{Pennebaker et~al.(2015)Pennebaker, Boyd, Jordan, and
  Blackburn}]{pennebaker2015a}
James~W. Pennebaker, Ryan~L. Boyd, Kayla Jordan, and Kate Blackburn. 2015.
\newblock The development and psychometric properties of liwc2015.
\newblock Technical report.

\bibitem[{Poria et~al.(2020)Poria, Hazarika, Majumder, and
  Mihalcea}]{poria2020a}
Soujanya Poria, Devamanyu Hazarika, Navonil Majumder, and Rada Mihalcea. 2020.
\newblock Beneath the tip of the iceberg: Current challenges and new directions
  in sentiment analysis research.
\newblock ArXiv:2005.00357 [cs], May. arXiv: 2005.00357.

\bibitem[{Sailunaz and Alhajj(2019)}]{sailunaz2019a}
Kashfia Sailunaz and Reda Alhajj. 2019.
\newblock Emotion and sentiment analysis from twitter text.
\newblock \emph{Journal of Computational Science}, 36(101003).

\bibitem[{Scott et~al.(2018)Scott, Sliwinski, Zawadzki, Stawski, Kim,
  Marcusson-Clavertz, Lanza, Conroy, Buxton, Almeida, and Smyth}]{scott2018a}
Stacey~B. Scott, Martin~J. Sliwinski, Matthew Zawadzki, Robert~S. Stawski,
  Jinhyuk Kim, David Marcusson-Clavertz, Stephanie~T. Lanza, David~E. Conroy,
  Orfeu Buxton, David~M. Almeida, and Joshua~M. Smyth. 2018.
\newblock A coordinated analysis of variance in affect in daily life.
\newblock \emph{Assessment:107319111879946}.

\bibitem[{Seyeditabari et~al.(2018)Seyeditabari, Tabari, and
  Zadrozny}]{seyeditabari2018a}
Armin Seyeditabari, Narges Tabari, and Wlodek Zadrozny. 2018.
\newblock Emotion Detection in Text: a Review. arXiv:1806.00674 [cs], June.
  arXiv: 1806.00674.

\bibitem[{Socher et~al.(2013)Socher, Perelygin, Wu, Chuang, Manning, Ng, and
  Potts}]{socher2013a}
Richard Socher, Alex Perelygin, Jean Wu, Jason Chuang, Christopher~D. Manning,
  Andrew Ng, and Christopher Potts. 2013.
\newblock Recursive deep models for semantic compositionality over a sentiment
  treebank.
\newblock In \emph{Proceedings of the 2013 Conference on Empirical Methods in
  Natural Language Processing}, pages 1631–1642,, Seattle, Washington, USA.
  Association for Computational Linguistics.

\bibitem[{van~der Vegt and Kleinberg(2020)}]{vegt2020a}
Isabelle van~der Vegt and Bennett Kleinberg. 2020.
\newblock Women worry about family, men about the economy: Gender differences
  in emotional responses to covid-19.
\newblock ArXiv:2004.08202 [cs], April. arXiv: 2004.08202.

\bibitem[{Westermann et~al.(1996)Westermann, Spies, Stahl, and
  Hesse}]{westermann1996a}
Rainer Westermann, Kordelia Spies, Günter Stahl, and Friedrich~W. Hesse. 1996.
\newblock Relative effectiveness and validity of mood induction procedures: a
  meta-analysis.
\newblock \emph{European Journal of Social Psychology}, 26(4).

\end{thebibliography}
\bibliographystyle{acl_natbib}

\end{document}